\documentclass[11pt,a4paper]{article}
\usepackage{acl2015}
\usepackage{times}
\usepackage{url}
\usepackage{latexsym}

\usepackage{amsmath}
\usepackage{multirow}
\usepackage{graphicx}
\usepackage{booktabs}
\usepackage{graphicx}
\usepackage{epstopdf}
\usepackage{caption}
\usepackage{subcaption}
\usepackage{amsmath}
\DeclareMathOperator*{\argmax}{arg\,max}

\usepackage{multicol}

\title{An Automatic Machine Translation Evaluation Metric Based on Dependency Parsing Model}

\author{Hui $\text{Yu}^\dag$\ \ \   Xiaofeng $\text{Wu}^\ddag$ \ \ \ Wenbin $\text{Jiang}^\dag$\ \ \  Qun $\text{Liu}^{\ddag\dag}$\ \ \ Shouxun $\text{Lin}^\dag$\\
 $^\dag$Key Laboratory of Intelligent Information Processing \\
 Institute of Computing Technology, Chinese Academy of Sciences  \\
 $^\ddag$ADAPT Centre, School of Computing, Dublin City University\\
 \\} 

\date{}

\begin{document}
\maketitle
\begin{abstract}
Most of the syntax-based metrics obtain the similarity by comparing the sub-structures extracted from the trees of hypothesis and reference. These sub-structures are defined by human and can't express all the information in the trees because of the limited length of sub-structures. In addition, the overlapped parts between these sub-structures are computed repeatedly. To avoid these problems, we propose a novel automatic evaluation metric based on dependency parsing model, with no need to define sub-structures by human. First, we train a dependency parsing model by the reference dependency tree. Then we generate the hypothesis dependency tree and the corresponding probability by the dependency parsing model. The quality of the hypothesis can be judged by this probability. In order to obtain the lexicon similarity, we also introduce the unigram F-score to the new metric. Experiment results show that the new metric gets the state-of-the-art performance on system level, and is comparable with METEOR on sentence level.

\end{abstract}
\section{Introduction}

Automatic machine translation (MT) evaluation  not only evaluates the performance of MT systems, but also accelerates the development of MT systems \cite{och2003minimum}. According to the type of the employed information, the automatic MT evaluation metrics can be classified into three categories, lexicon-based metrics, syntax-based metrics and semantic-based metrics.

Most of the syntax-based evaluation metrics obtain the similarity between reference and hypothesis by comparing the sub-structures between the trees of reference and hypothesis, such as HWCM \cite{Liu05syntacticfeatures} and the LFG-based metric \cite{Owczarzak:2007:DAE:1626281.1626292}. 
HWCM uses the headword chains extracted from the dependency tree, while the LFG-based metric uses the Lexical-Functional Grammar dependency tree. 
Some syntax-based metrics calculate the similarity between the sub-structure of the reference tree and the string of hypothesis, such as BLEU$\widehat{A}$TRE \cite{mdobrink:BLEUATRE} and RED \cite{yu-EtAl:2014:Coling3}.
The sub-structures in these metrics are defined by human and can't express all the information in the trees because of the limited length of sub-structures. In addition, the overlapped parts between these sub-structures are computed repeatedly. 

To avoid the above defects, we propose a new metric from the view of dependency tree generation. We don't need to define sub-structures by human for the new metric. We train a dependency parsing model by the reference dependency tree. By this model, we can obtain the dependency tree of the hypothesis and the corresponding probability which is also the score of the dependency parsing model. 
The syntactic similarity between the hypothesis and the reference can be judged by this score. In order to obtain the lexicon similarity, we also introduce the unigram F-score to the new metric. 
The experiment results show that the new metric gets the-state-of-art performance on system level evaluation, and gets the comparable correlation with METEOR on sentence level evaluation.

The remainder of this paper is organized as follows. Section 2 describes the maximum-entropy-based dependency parsing model. Section 3 presents the new  MT evaluation metric based on dependency parsing model. Section 4 gives the experiment results. Conclusions and future work are discussed in Section 5.

\section{Maximum-entropy-based Dependency Parsing Model }

Shift-reduce algorithm is used in the dependency parsing model. In the shift-reduce algorithm, the input sentence is scanned from left to right. In each step, one of the following two actions is selected, shift the current word into the stack or reduce the two (or more than two) items on the top of the stack to one item. 

Generally, the reduce action includes two sub-actions $reduce_L$ and $reduce_R$. $reduce_L$ means that the left item is considered as the head after reducing, and $reduce_R$ means that the right item is considered as the head after reducing. 
Formally, the transition state in the shift-reduce parser can be represented as a tuple $<S, Q, A>$. S is a stack. 
Q is a sequence of unprocessed words. A is the already-built set of dependency arcs, which is part of the dependency tree in the current state. In each step, one of the following three actions is selected.
\begin{itemize}
\item $shift$: shift the head word in the queue Q into the Stack S.
\item $reduce_L$: merge the top two items ($s_t$ and $s_{t-1}$) in S into $s_t$, $t>=2$. $s_t$ is considered as the head, and the left arc ($s_t$, $s_{t-1}$) is added to the set A.
\item $reduce_R$: merge the top two items ($s_t$ and $s_{t-1}$)  in S into $s_{t-1}$, $t>=2$. $s_{t-1}$is considered as the head, and the right arc ($s_{t-1}$, $s_t$) is added to the set A.
\end{itemize}

In the traditional shift-reduce decoder algorithm, the next  action can be predicted by Formula \eqref{pre-action}, when the state of the dependency parser is $s$. In Formula \eqref{pre-action}, $action=\{shift,reduce_L, reduce_R\}$. $score_{act}(T,s)$ is the score of action $T$ when the current state is $s$.
\begin{equation}
T(s)=argmax _{T\in action} score_{act}(T,s)
\label{pre-action}
\end{equation}

We use the method of classification to decide which action should be chosen in the transition sequence. 
We combine the action and the corresponding context as a training example, which describes which action should be chosen in a certain context. The context can be represented as a series of features. 
The feature templates used in this paper are the same as those used in Huang et al. \shortcite{huang2009bilingually}. 

We use the maximum entropy as the classification method to train the examples and get $model_{ME}$. 
When calculating the score of a transition action, we use Formula \eqref{3-line}.
\begin{equation}
score_{act}(T',s)= \sum_i \lambda_i f_i(T',s)
\label{3-line}
\end{equation}
$f_i(T',s)$ is the i$th$ feature when the current state is $s$ and the transition action is $T'$. $\lambda _i$ is the weight of the i$th$ feature. In shift-reduce algorithm, there are three kinds of actions in  each transition action. 
The probability that the scores of all the three actions are zero is very low, because the feature templates include POS (Part-of-Speech) of the current word and POS of the two words before the current word. 
If $model_{ME}$ chooses two kinds of actions, the score of the third action is zero. 
To avoid  the zero score, we use the normalization method in Formula \eqref{3-regular}. $P_{act}(T',s)$ is the normalized probability of the chosen action $T'$ when the current state is $s$. $z$ is the constant for normalization. $set(s)$ in Formula \eqref{3-z} is the set of all possible actions when the current state is $s$. 
 
\begin{equation}
P_{act}(T',s)=\frac{1}{z} \cdot exp(\sum_i \lambda_i f_i(T',s))
\label{3-regular}
\end{equation}

\begin{equation}
z=\sum_{T'\in set(s)} exp( \sum_i \lambda_i f_i(T',s))
\label{3-z}
\end{equation}

Beam search algorithm \cite{zhang2008tale} is used in shift-reduce decoder algorithm. 
For a sentence $x$, we can get many dependency trees and use $gen(x)$ to represent the set of the dependency trees. 
Then the best one can be obtained by Formula \eqref{formula-parser}. 
$actset(y)$ represents the set of all the actions when generating dependency tree $y$.

\begin{equation}
tree(x)=\argmax _{y\in gen(x)} \sum _{T' \in actset(y)} \log(P_{act}(T', s_{T'}))
\label{formula-parser}
\end{equation}

$model_{ME}$ is trained with the data which contain the information in the process of dependency parsing and is used to parse a sentence. 
So we name the trained model $model_{ME}$ as dependency parsing model. 
The score of the dependency parsing model is defined in Formula \eqref{formula-dep}.
\begin{equation}
Score(x)=\sum _{T' \in actset(tree(x))} \log(P_{act}(T', s_{T'}))
\label{formula-dep}
\end{equation} 

\section{Dependency-parsing-model-based MT Evaluation Metric}

\subsection{Training of Dependency Parsing Model}
We should get the reference dependency tree first for training dependency parsing model. The reference dependency tree can be generated by the open-source tools or labeled by human. We use the Stanford tools\footnote{\url{http://nlp.stanford.edu/software/stanford-dependencies.shtml}} to generate reference dependency tree.
After obtaining the reference dependency tree, we can use it to train the dependency parsing model. The reference dependency tree is used as training corpus to extract features, according to the feature templates defined in Huang et al. \shortcite{huang2009bilingually}. 
A training example is achieved by combining the features and the action in shift-reduce algorithm. The format of the training example is shown in Table \ref{feat-exam}. 
\begin{table}[h]
\centering
\begin{tabular}{|l|l|}
\hline
Action & Features(context) \\
\hline
SHIFT &s0w-s0t=Economia $\mid$ NNP s0w=...... \\
\hline
RIGHT &s0w-s0t=and $\mid$ CC s0w=and  ...... \\
\hline
LEFT &s0w-s0t=link $\mid$ VB s0w=link ...... \\
\hline
......&...... \\
\hline
\end{tabular}
\caption{ The format of training example. s0w represents the word on the top of stack. s0t represents the POS of the top word in stack.
}
\label{feat-exam}
\end{table}
We train the extracted examples using the maximum entropy and get a dependency parsing model. According to the method introduced in Section 2, we parse the hypothesis using this dependency parsing model. We can get a Score(hyp) of the dependency parsing model for hypothesis hyp as in Formula \eqref{formula-dep}.

We train a dependency parsing model for each sentence separately. That is to say, the reference dependency tree of sentence $i$ is only used to train the dependency parsing  model for the hypothesis of sentence $i$. 
We also tried other methods, such as using all the reference dependency trees to train the model for each hypothesis, or adding a background corpus together with the reference dependency tree to train the model for each hypothesis. For the above two methods, we give a higher weight to the dependency tree of sentence $i$ when training the model for hypothesis $i$. However, for these two methods, the performance is worse than only using the reference dependency tree of sentence $i$ when training the model for hypothesis $i$.

The dependency parsing model is trained by maximum entropy model, which can ensure smoothness when satisfying all of the conditions. 
In the case of data sparse, all the features of all the actions in a state may be zero, according to Formula \eqref{3-regular}. For this state, the probabilities of all the actions are equal.
Sometimes none of the words in hypothesis appears in reference, but the POS of some words may appear in the reference. The dependency parsing model can differentiate this case, because  POS is used in the feature templates. Table \ref{3-POS} gives a reference, two hypotheses and the corresponding POS sequences of the three sentences. We can see that, none of the words in hyp1 or hyp2 appears in the reference, but the POS of some words appear in the reference. According to the dependency parsing model defined in Formula \eqref{formula-dep},  we can get $Score(hyp1)=-4.46$ and $Score(hyp2)=-5.87$. From these two scores, we can conclude that hyp1 is better than hyp2, which is the truth.

\begin{table*}[!ht]
\centering
\begin{tabular}{|l|l|l|}
\hline
& \textbf{word sequence} & \textbf{POS sequence}\\
\hline
ref & my objective is to discover the truth . & PRP NN VBZ TO VB DT NN . \\
\hline
hyp1 & our goal was finding fact . & PRP NN VBZ VBG NN . \\
\hline
hyp2 & was finding our goal fact . &VBZ VBG PRP NN NN .\\
\hline
\end{tabular}
\caption{An example for the case that none of the words in hyp1 or hyp2  appears in reference but the POS of some words appear in the reference. }
\label{3-POS}
\end{table*}

\subsection{Normalization of the Dependency Parsing Model Score }
A transition sequence is obtained in the process of generating the dependency tree according to the shift-reduce algorithm. Each word in the sentence should be pushed into the stack once, and each word is popped from the stack once for reduction except the root node. Therefore, there are $n$ steps of shift actions and $n-1$ steps of reduce actions, $2n-1$ actions in all, which means that the length of the transition sequence is $2n-1$. $n$ is the length of the sentence. 
The score of the dependency parsing model is the sum of the logarithms of the transition actions' probabilities, 
as in Formula \eqref{formula-dep}. 
Because the value is negative after the logarithm, it will cause penalty for long sentences. Some sentences can achieve high scores because of a shorter length and not because of higher quality. Therefore, we need to normalize the score of the dependency parsing model, as in Formula \eqref{prode-norm}. $hyp$ is a hypothesis. $n$ is the length of $hyp$. $Score(hyp)$ is defined in Formula \eqref{formula-dep}.
The normalized score of the Dependency Parsing Model is named as DPM which is a value between 0 and 1. 

\begin{equation}
DPM= \exp(\frac{Score(hyp)}{2n -1}  )\\
\label{prode-norm}
\end{equation}

\subsection{Lexical Similarity}
Dependency parsing model mainly evaluates the syntax structure similarity between the reference and the hypothesis. Besides the syntax structure, another important factor is the lexical similarity. Therefore, unigram F-score is used to represent the lexical similarity in our metric.

F-score can be calculated by Formula \eqref{fscore}. $\alpha$ is a decimal between 0 and 1, which can balance the effects of precision and recall. $P$ means precision and $R$ means recall.
\begin{equation}
\textit{F-score}=\dfrac{P \times R}{\alpha \times P + (1-\alpha) \times R}
\label{fscore}
\end{equation}

Many automatic evaluation metrics can only find the exact match between the reference and the hypothesis, and the information provided by the limited number of references is not sufficient. Some evaluation metrics, such as TERp \cite{snover2009fluency} and METOER \cite{banerjee-lavie:2005:MTSumm,lavie2009meteor,denkowski:lavie:meteor-wmt:2014}, introduce extra resources to expand the reference information. We also introduce some extra resources when calculating F-score, such as stem \cite{porter2001snowball}, synonym\footnote{http://wordnet.princeton.edu} and paraphrase. 
First, we obtain the alignment with Meteor Aligner \cite{denkowski:lavie:meteor-wmt:2011} in which exact, stem, synonym and paraphrase are all considered. Then we can find the matched words using the alignment, and every matched word corresponds to a match module type (exact, stem, synonym or paraphrase). Different match module types have different match weights, which can be represented as $w_{exact}$, $w_{stem}$, $w_{synonym}$ and $w_{paraphrase}$.  

The words within a sentence can be classified into content words and function words. The effects of the two kinds of words are different and they should not have the same matching score, so we introduce a parameter $w_f$ to distinguish them. 

After introducing extra resources, the precision $P$ and recall $R$ can be calculated by Formula \eqref{precision} and Formula \eqref{recall} respectively.
\begin{equation}
P=\frac{\sum _i m_i \cdot (  w_{f} \cdot f_h(i) +  (1- w_f) \cdot c_h(i))}{w_f \cdot num_c(h) + (1-w_f) \cdot num_f(h)} \\
\label{precision}
\end{equation}
\begin{equation}
R=\frac{\sum _i m_i \cdot (  w_{f} \cdot f_r(i) +  (1- w_f) \cdot c_r(i))}{w_f \cdot num_c(r) + (1-w_f) \cdot num_f(r)} \\
\label{recall}
\end{equation}

In Formula \eqref{precision}, $i$ is the i$th$ word in the matched unigrams, $0<i \leq n$, and $n$ is the number of the matched unigrams. $m_i$ is the weight of the match module which the i$th$ matched word belongs to. $w_f$ is the weight of function words. $num_f(h)$ is the number of function words in the hypothesis, and $num_c(h)$ is the number of content words in the hypothesis. $f_h(i)$ represents whether the i$th$ matched unigram in hypothesis is function word. 
\begin{equation*}
f_h(i)=\left\{
\begin{aligned}
 1&  & if\ function\ word \\
0&  & if\ not\ function\ word
\end{aligned}
\right.
\label{func-h}
\end{equation*}
$c_h(i)$ represent whether the i$th$ matched unigram in hypothesis is content word. 
\begin{equation*}
c_h(i)=\left\{
\begin{aligned}
 1&  & if\ content\ word \\
0&  & if\ not\ content\ word
\end{aligned}
\right.
\label{func-h}
\end{equation*}

In Formula \eqref{recall}, $i$, $m_i$ and $w_f$ have the same meanings as those in Formula \eqref{precision}. $num_f(r)$ and $num_c(r)$ are the number of function words and content words respectively in reference. $f_r(i)$ represents whether the i$th$ matched word in reference is function word. 
\begin{equation*}
f_r(i)=\left\{
\begin{aligned}
 1&  & if\ function\ word \\
0&  & if\ not\ function\ word
\end{aligned}
\right.
\label{func-h}
\end{equation*}
$c_r(i)$ represent whether the i$th$ matched unigram in reference is content word. 
\begin{equation*}
c_r(i)=\left\{
\begin{aligned}
 1&  & if\ function\ word \\
0&  & if\ not\ function\ word
\end{aligned}
\right.
\label{func-h}
\end{equation*}

\begin{table*}[!ht]
\centering
\begin{tabular}{|l|l|}
\hline
Parameter & Meaning  \\
\hline
$\alpha$ & balance the effects of precision and recall \\
\hline
$w_{f}$ & differentiate the effects of function word and content word\\
\hline
$w_{exact}$ & match weight for match module type \textit{exact}\\
\hline
$w_{stem}$ & match weight for match module type \textit{stem}\\
\hline
$w_{synonym}$ & match weight for match module type \textit{synonym} \\
\hline
$w_{paraphrase}$ &  match weight for match module type \textit{paraphrase}\\
\hline
\end{tabular}
\caption{The meanings of parameters in DPMF.
}
\label{para-meaning}
\end{table*}

\begin{table*}[!ht]
\centering
\begin{tabular}{|c|c|c|c|c|c|c|}
\hline
data& cs-en &de-en &es-en &fr-en &ru-en &hi-en\\
\hline
WMT2012 & 6 & 16 & 12 & 15 & - &-\\
\hline
WMT2013& 12 & 23 & 17 & 19 &23&- \\
\hline
WMT2014 &5 & 13 &-& 8& 13&9 \\
\hline
\end{tabular}
\caption{The number of translation systems for each language pair on WMT 2012, WMT 2013 and WMT 2014. cs-en means Czech to English. de-en means German to English. es-en means Spanish to English. fr-en means French to English. ru-en means Russian to English. hi-en means Hindi to English.}
\label{data-into}
\end{table*}

\begin{table*}[!ht]
\centering
\begin{tabular}{|c|c|c|c|c|c|c|}
\hline
language pair& $\alpha$ & $w_f$& $w_{exact}$ & $w_{stem}$ & $w_{synonym}$ &$w_{paraphrase}$   \\
\hline
*-en & 0.85 & 0.25  &1.0 &0.6 &0.8 &0.6  \\
\hline
\end{tabular}
\caption{Parameter values of DPMF.  *-en represents all the language pairs with English as target language.
}
\label{value-into}
\end{table*}

\subsection{Final Score of DPMF}
After obtaining the score of dependency parsing model and lexical similarity, we can calculate the final score of the new metric. Because we use both the Dependency Parsing Model and F-score, we name the score as DPMF. As in Formula \eqref{prode-last}, DPMF can evaluate the similarities both on syntax and on lexicon.
\begin{equation}
\begin{split}
DPMF&=DPM\times \emph{F-score} \\
\end{split}
\label{prode-last}
\end{equation}

The system level score is the average score of all the sentences. 
There are some parameters when calculating F-score. The meaning of each parameter is listed in Table \ref{para-meaning}.

\section{Experiment}

To verify the effectiveness of DPM and DPMF, we carry out experiments on both the system level and the sentence level. \footnote{Interested readers can find the source code of DPM and DPMF from https://github.com/YuHui0117/AMTE/tree/master/DPMF.}

\subsection{Data}
The data used in the experiment are WMT 2012, WMT 2013 and WMT 2014. The language pairs are Czech-to-English, German-to-English, Spanish-to-English, French-to-English, Russian-to-English and Hindi-to-English. The number of translation systems for each language pair are shown in Table \ref{data-into}.

All the parameters of DPMF are also included in METEOR and METEOR has tuned these parameters for better performance. 
So we use the same parameter values as METEOR as empirical value in DPMF and don't need to tune the parameters again. The parameter values used in the experiment are listed in Table \ref{value-into}.

\subsection{System Level Correlation}

\begin{table*}[!ht]
\begin{subtable}{\textwidth}
\centering
\begin{tabular}{|l|c|c|c|c|c|}
\hline
 metrics& \ \ cs-en \ \  & \ \ de-en\ \   &\ \ es-en \ \  & \ \ fr-en \ \ &avg \\
\hline 
TER& .886           & .624          &.916           & .821          & .812 \\
\hline
BLEU & .886         &.671           &.874           &.811           &.811  \\
\hline
METEOR& .657        & .885          & .951          &.843           &.834\\
\hline
$\bullet$SEMPOS \ \ \ \ \ \ \ \ \ \ \ \ \ \ \  \ \ \ \ \ \ \ \ \ \  \ \ \ \ & .940       & \textbf{.920} & .940          & .800          & .900\\
\hline
DPM            &.943 &.735 &.888 &.821 &.847 \\
DPMF &\textbf{.943}& .909          & \textbf{.951} & \textbf{.850} & \textbf{.913} \\
\hline
\end{tabular}
\caption{System level correlations on WMT2012.
}
\label{3-rst-sys-wmt12}
\end{subtable}

\begin{subtable}{\textwidth}
\centering
\begin{tabular}{|l|c|c|c|c|c|c|}
\hline
 metrics&  cs-en & de-en &es-en & fr-en &ru-en & avg \\
\hline
TER     &   .800        & .833          &.825           &.951           &.581           &.798 \\
\hline
BLEU    &  .946         & .851          & .902	         & .\textbf{989} 	    &.698           &	.877 \\
\hline
$\bullet$METEOR  \ \ \ \ \ \ \ \ \ \ \ \ \ \ \ \ \ \ \ \ \ \ \ \  \ \ \ \ \ \ &   .964        & .961          & .979 &	.984       & .789             &	.935\\
\hline
DPM        &.945 &.880 &.937 &.951 &.800 &.903 \\
DPMF     &\textbf{.991} &\textbf{.975} &.\textbf{993} &.984 &\textbf{.849} &\textbf{.958} \\
\hline
\end{tabular}
\caption{System level correlations on WMT2013.
}
\label{3-rst-sys-wmt13}
\end{subtable}

\begin{subtable}{\textwidth}
\centering
\begin{tabular}{|l|c|c|c|c|c|c|}
\hline
 metrics&  cs-en & de-en &fr-en & hi-en & ru-en & avg \\
\hline
TER                 & .976          & .775          & .952          & .618          &  .809         & .826 \\
\hline
BLEU                & .909           & .832         & .952          & .956 &  .789         &  .888 \\
\hline
METEOR              & .980          &  .927         &  .975         &.457           &   .805        & .829 \\
\hline
\hline
$\bullet$*DISCOTK-PARTY-TUNED &.975 &\textbf{.943} & \textbf{.977} & .956 & \textbf{.870} & \textbf{.944} \\
*LAYERED             &.941 &.893 &.973 &\textbf{.976} &.854 & .927 \\
*DISCOTK-PARTY       &.983 &.921 &.970 &.862 &.856&.918 \\
*UPC-STOUT           &.948 &.915 &.968 &.898 &.837 &.913 \\
\hline
\hline
VERTA-W          &.934 &.867 &.959 &.920 &.848 &.906 \\
\hline
DPM        &.988 &.817 &.946 &.934 &.858 &.909 \\
DPMF  & \textbf{.999}  & .920         & .967          & .882          & .832          & .920 \\
\hline
\end{tabular}
\caption{System level correlations on WMT2014.
}
\label{3-rst-sys-wmt14}
\end{subtable}
\caption{System level correlations on WMT 2012, WMT 2013 and WMT 2014. The value in bold is the best result in each column.  \textit{avg} stands for the average result of all the language pairs for each metric on WMT 2012, WMT 2013 or WMT 2014. Metrics with * are the hybrid metrics. Metrics with $\bullet$ are the best performance metrics in each data set. }
\label{3-rst-sys}
\end{table*}

To evaluate the correlation with human judges, Spearman's rank correlation coefficient $\rho$ is used for system level. $\rho$ is calculated using Formula \eqref{rho}. 
\begin{equation}
\rho =1-\dfrac{6\sum d_i^2}{n(n^2-1)}
\label{rho}
\end{equation}
$d_i$ is the difference between the human rank and metric’s rank for system $i$. $n$ is the number of systems.


In the experiment, we give the correlations of DPM and DPMF respectively. For comparison, the baseline metrics are the widely-used metrics, BLEU\footnote{ftp://jaguar.ncsl.nist.gov/mt/resources/mteval-v13a.pl}, TER\footnote{http://www.cs.umd.edu/~snover/tercom}  and METEOR\footnote{http://www.cs.cmu.edu/~alavie/METEOR/download/meteor-1.4.tgz}.
In addition, we also give the correlations of the metrics with the best performance on average according to the published results of WMT 2012, WMT 2013 and WMT 2014. 
For WMT 2012 and WMT 2013, the metrics with the best performance on average are SEMPOS \cite{machavcek2011approximating} and METEOR respectively. 
For WMT 2014, the top-four metrics 
are DISCOTK-PARTY-TUNED \cite{joty-EtAl:2014:W14-33}, LAYERED \cite{gautam-bhattacharyya:2014:W14-33}, DISCOTK-PARTY \cite{joty-EtAl:2014:W14-33} and UPC-STOUT \cite{gonzalez-barroncedeno-marquez:2014:W14-33}. They are all hybrid metrics\footnote{Hybrid metrics directly use the scores of many kinds of metrics, such as BLEU, TER, METEOR and some syntax-based metrics, so we think they are hybrid metrics. 
For the metrics using different kinds of information types (lexicon, syntax and semantic information) as features, we still think they are single metrics, because they don't use the score of other metrics.} which include many kinds of other metrics.
For fairness, we also give the result of the metric with the best performance on average in the single metrics, VERTA-W \cite{comelles-atserias:2014:W14-33}.

\begin{table*}[!ht]
\begin{subtable}{\textwidth}
\centering
\begin{tabular}{|l|c|c|c|c|c|}
\hline
Language& \ \ cs-en \ \  & \ \ de-en\ \  &\ \  es-en\ \  &\ \  fr-en\ \  &avg   \\
\hline
BLEU 		 &.157 			& .191			& .189			&.210         & .187 \\
\hline
METEOR		 & .212			&.275 			&.249 			&.251          &.247  \\
\hline
$\bullet$spede07\_pP  \ \ \  \ \ \ \ \ \ \ \ \ \ \ \ \ \ \ \ \ \ \ \ \ \ \ \ \ &.212 			&.278 			&.265 			&\textbf{.260} &.254 \\
\hline
DPM           &.146 			&.187 			&.211 			&.183          &.182 \\
DPMF    &\textbf{.227} &\textbf{.279}  &\textbf{.279}  &.252          &\textbf{.259} \\
\hline
\end{tabular}
\caption{Sentence level correlations on WMT 2012.
}
\label{3-rst-sent-wmt12}
\end{subtable}
\begin{subtable}{\textwidth}
\centering
\begin{tabular}{|l|c|c|c|c|c|c|}
\hline
Language &cs-en & de-en &es-en & fr-en &ru-en & avg \\
\hline
BLEU  & .199            & .220          & .259          & .224          & .162          & .213 \\
\hline
METEOR & \textbf{.265}	& .293&  .324	&.264& \textbf{.239}              &.277 \\
\hline
$\bullet$SIMPBLEU-RECALL  \ \ \ \ \ \ \ \ \ \ &.260 &\textbf{.318} &\textbf{.387} &\textbf{.303} &.234 &\textbf{.301} \\
\hline
DPM           &.179 &.204 &.237 &.194 &.146 &.192 \\
DPMF & .258           & .296          & .316          & .269          & .227 & .273 \\
\hline
\end{tabular}
\caption{Sentence level correlations on WMT 2013.
}
\label{3-rst-sent-wmt13}
\end{subtable}
\begin{subtable}{\textwidth}
\centering
\begin{tabular}{|l|c|c|c|c|c|c|}
\hline
Language &cs-en & de-en &fr-en & hi-en & ru-en & avg \\
\hline
BLEU  &.216             & .259          & .367          & .286          &  .256         & .277   \\
\hline
METEOR &  .282   &  .334 &  .406 & .420    &  .329 &  .354  \\
\hline
BEER &.284 &.337 &.417 &\textbf{.438} &.333 &.362 \\
\hline
$\bullet$*DISCOTK-PARTY-TUNED &\textbf{.328} &\textbf{.380} &\textbf{.433} &.434 &\textbf{.355} &\textbf{.386}\\
\hline
DPM        &.182 &.224 &.331 &.301 &.243 &.256 \\
DPMF&.283              & .332         & .404          &.426 & .324          & .354 \\
\hline
\end{tabular}
\caption{Sentence level correlations on WMT 2014.
}
\label{3-rst-sent-wmt14}
\end{subtable}
\caption{Sentence level correlations on WMT 2012, WMT 2013 and WMT 2014. The value in bold is the best result in each column.  \textit{avg} stands for the average result of all the language pairs for each metric on WMT 2012, WMT 2013 or WMT 2014. Metrics with * are the hybrid metrics. Metrics with $\bullet$ are the best performance metrics in each data set.}
\label{3-rst-sent}
\end{table*}

System level correlations are shown in Table \ref{3-rst-sys}. 
According to Table \ref{3-rst-sys}, DPM can get higher correlations than BLEU and TER on the three data sets. 
DPM also gets higher correlations than METEOR on WMT 2012 and WMT 2014. The experiment results show that DPM can effectively evaluate the hypothesis. 
In order to evaluate the lexical information, we also introduce the F-score to DPM and add some extra linguistic resources to F-score to more accurately evaluate the similarity between the hypothesis and the reference on lexicon. 
After adding F-score, the performance of DPMF is greatly  improved over DPM on the three data sets. 
So it is effective to add F-score to DPM to evaluate the lexical information. On WMT 2012, WMT 2013 and WMT 2014, DPMF gets higher correlations than METEOR. Compared with the best metric SEMPOS in WMT 2012, DPMF achieves higher correlations on the three language pairs cs-en, es-en and fr-en, and gets 1.3 points improvement over SEMPOS on average. 
Compared with the best metric METEOR in WMT 2013, DPMF achieves higher correlations on all the language pairs except an equal correlation on fr-en. On average, DPMF obtains 2.3 points improvement over METEOR. Compared with the best single metric VERTA-W in WMT 2014, the correlation improvement of DPMF is 1.4 points. 
DPMF also outperforms the hybrid metrics LAYERED and DISCOTK-PARTY, but there is still some work to do to catch up with the best hybrid metric for DPMF.

\subsection{Sentence Level Correlation}

To evaluate the performance of DPM and DPMF further, we also carry out the experiments on sentence level.
On sentence level, Kendall's  $\tau$  correlation coefficient is used.
$\tau$ is calculated using the following equation.
\begin{equation*}
\tau = \dfrac{\text{num\_con\_pairs} - \text{num\_dis\_pairs}}{\text{num\_con\_pairs} + \text{num\_dis\_pairs}}
\end{equation*}
$num\_con\_pairs$ is the number of concordant pairs and $num\_dis\_pairs$ is the number of disconcordant pairs.


In the experiments, we give the results of DPM and DPMF respectively. For comparison, the baseline metrics are the widely-used metrics, BLEU and METEOR.
In addition, we also give the correlations of the metric with the best performance on average according to the published results of WMT 2012, WMT 2013 and WMT 2014. 
The metrics with the best performance on average are spede07\_pP on WMT 2012, SIMPBLEU-RECALL on WMT 2013 and DISCOTK-PARTY-TUNED on WMT 2014 respectively. Because DISCOTK-PARTY-TUNED is a hybrid metric, we also give the result of the single metric with the best performance on average, BEER \cite{stanojevic-simaan:2014:W14-33}.

Sentence level correlations are shown in Table \ref{3-rst-sent}.
From Table \ref{3-rst-sent}, we can see that the performance of DPM is not good and a little lower than BLEU. The reason is that DPM mainly considers the syntactic structure information. After introducing lexical information (F-score), DPMF achieves a significant improvement over DPM and BLEU. DPMF outperforms METEOR on WMT 2012 and is comparable with METEOR on WMT 2013 and WMT 2014. The above results show that DPMF can give an effective evaluation for the hypothesis on sentence level. Compared with the best metric spede07\_pP on WMT 2012, DPMF can achieve a comparable correlation. 

\section{Conclusion and Future Work}
In this paper, we propose a novel dependency-parsing-model-based automatic MT evaluation metric DPMF. 
DPMF evaluates the syntactic similarity through the score of hypothesis dependency parsing model and evaluates the lexical similarity by unigram F-score.
The syntactic similarity method is designed from the view of dependency tree generation, which is totally different from the method of comparing the sub-structures and avoids the defects of defining sub-structures by human.
The experiment results show the effectiveness of DPMF on both system level evaluation and sentence level evaluation. DPMF gets the-state-of-art performance on system level on WMT 2012, WMT 2013 and WMT 2014. On sentence level, the performance of DPMF is comparable with METEOR on all of the three data sets.

In future, we will continue our work in two directions. 
When generating the hypothesis dependency tree, the model is trained  only using a limited number of reference sentences (only one reference for WMT corpus), so one direction is that we will enrich the references. The other direction is that we will apply DPMF to the tuning process of statistical machine translation to improve the translation quality.



\bibliographystyle{acl}
\bibliography{mte}

\end{document}